\documentclass[letterpaper, 10 pt, journal, twoside]{IEEEtran}
\usepackage{booktabs}
\usepackage{multirow}
\usepackage{xcolor}
\usepackage{caption}
\usepackage{gensymb}
\usepackage{url}
\usepackage{epsfig} 
\usepackage{mathptmx} 
\usepackage{times} 
\usepackage{amsmath} 
\usepackage{amssymb}  
\usepackage{subfigure}
\usepackage{array}
\usepackage{color, soul}
\usepackage[colorlinks,linkcolor=red,anchorcolor=blue,citecolor=green]{hyperref}
\usepackage{booktabs}
\usepackage{colortbl}
\usepackage{tcolorbox}
\usepackage{color,xcolor}
\usepackage{multirow}

\usepackage{float}
\usepackage{cite}
\usepackage{xcolor}
\usepackage{algpseudocode}
\usepackage[export]{adjustbox}
\usepackage{balance}
\usepackage{rotating}
\usepackage{graphicx}

\usepackage[utf8]{inputenc}

\title{\LARGE \bf
PillarTrack: Boosting Pillar Representation for Transformer-based \\ 3D Single Object Tracking on Point Clouds}

\author{Weisheng Xu$^{1}$$^\dagger$, Sifan Zhou$^{2}$$^\dagger$*, Jiaqi Xiong$^{3}$, Ziyu Zhao$^{2}$, Zhihang Yuan$^{4}$
\thanks{$^{1}$Weisheng Xu is with the School of Mathematics and Computer Science,
        Nanchang University, Nanchang 330031, China.
        {\tt\small wesleyxu224@gmail.com}, $^{2}$Sifan Zhou and Ziyu Zhao is with the School of Automation, Southeast University, Nanjing 210096, China.
        {\tt\small sifanjay@gmail.com}, $^{3}$Jiaqi Xiong is with Aberdeen Institute of Data Science and Artificial Intelligence, South China Normal University,
        Guangzhou, 528225, China. $^{4}$Zhihang Yuan is with Houmo AI,
        Beijing 100088, China.}
\thanks{$^\dagger$ Authors with equal contribution. * Corresponding author.}
}
\begin{document}
\maketitle
\begin{abstract}
LiDAR-based 3D single object tracking (3D SOT) is a critical issue in robotics and autonomous driving. Existing 3D SOT methods typically adhere to a point-based processing pipeline, wherein the re-sampling operation invariably leads to either redundant or missing information, thereby impacting performance. To address these issues, we propose PillarTrack, a novel pillar-based 3D SOT framework. First, we transform sparse point clouds into dense pillars to preserve the local and global geometrics. Second, we propose a Pyramid-Encoded Pillar Feature Encoder (PE-PFE) design to enhance the robustness of pillar feature for translation/rotation/scale. Third, we present an efficient Transformer-based backbone from the perspective of modality differences. Finally, we construct our PillarTrack based above designs. Extensive experiments show that our method achieves comparable performance on the KITTI and NuScenes datasets, significantly enhancing the performance of the baseline. 

\end{abstract}
\vspace{-2mm}
\section{INTRODUCTION}  

3D single object tracking (3D SOT) has a wide applications in robotics, such as a robot needs to track their master to complete material handling tasks in factories. Given the initial state (size and location) of target in the first frame, 3D single object tracking aims to estimate its 3D state across subsequent frames. Existing LiDAR-based 3D single object tracking methods \cite{fang20203d,p2b,ptt,bat,lttr,zhou2022pttr,hui2022stnet} generally follow the Siamese paradigm from 2D visual object tracking~\cite{SiamFC,SiameseRPN}, aiming to achieve a trade-off between runtime and accuracy. Given the sparse and irregular input point cloud, these methods initially leverage PointNet families~\cite{qi2017pointnet, qi2017pointnet++} to learn point-wise representation, and then obtain point-wise similarity with a similarity module. Finally, they estimate the state of specific targets based on these similarity features. SC3D \cite{sc3d} stands as the pioneering LiDAR-based 3D Siamese tracker, leveraging an efficient auto-encoder based on PointNet\cite{qi2017pointnet} for point-wise feature encoding. Subsequently, approaches such as P2B~\cite{p2b}, 3D-SiamRPN\cite{fang20203d}, BAT~\cite{bat} and PTT~\cite{ptt,ptt-journal} employ PointNet++~\cite{qi2017pointnet++} to extract point-wise representation and achieve superior tracking performance. However, one common issue with above point-based 3D SOT methods is the requirement for re-sampling the input points to a fixed number. For instance. P2B~\cite{p2b} re-samples the input points from 1024 to 512. This re-sample operation inevitably introduces the potential for redundant or lost information, which can adversely affect performance.  
\begin{figure}[t] 
  \centering
  \includegraphics[width=0.99\linewidth]{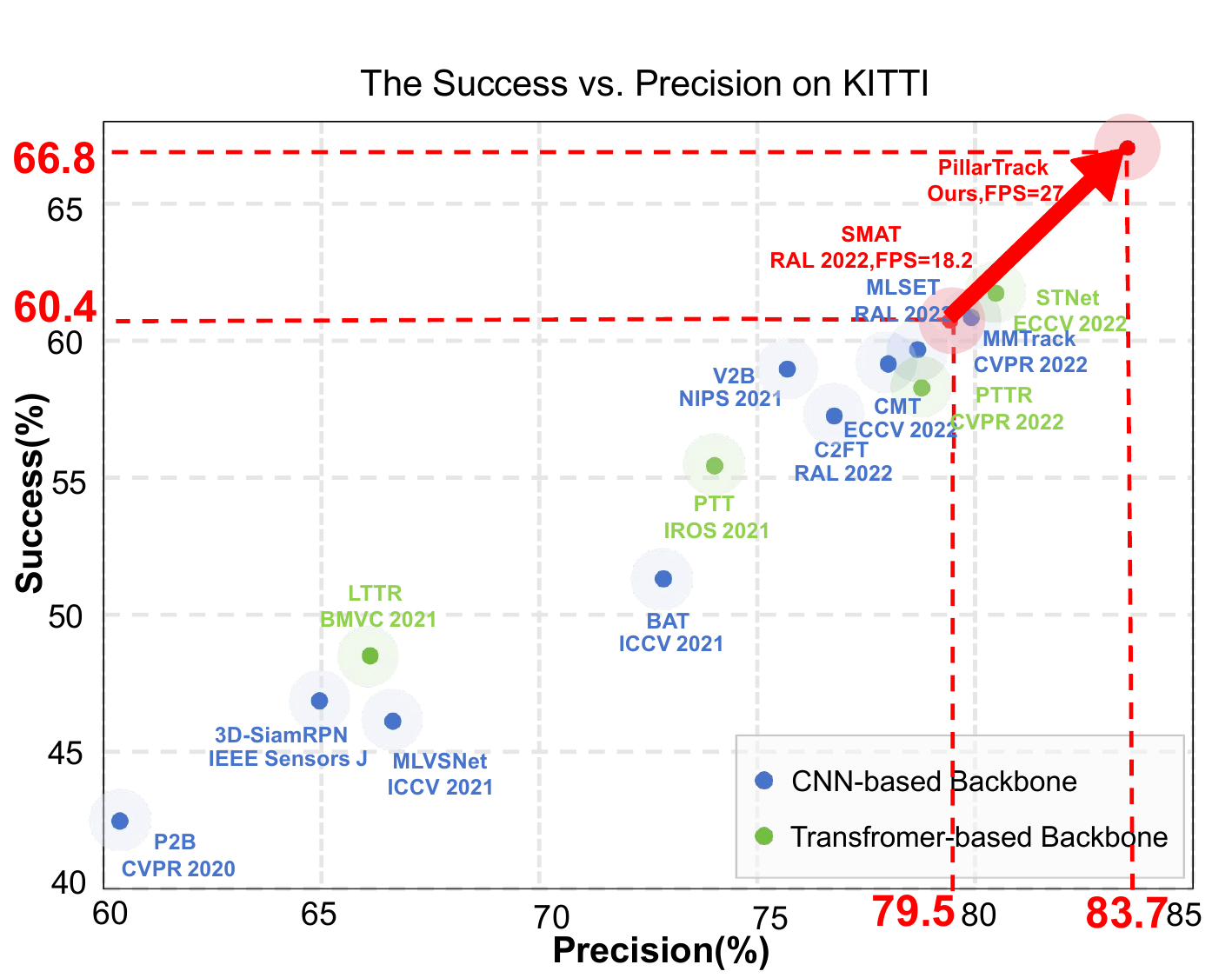}\vspace{-2mm}
  \caption{Comparison with other 3D SOT methods on KITTI dataset. We classify methods based on the backbone architecture and report performance on Success and Precision.}
  \label{performance}
  \vspace{-6mm}
\end{figure}

To this end, inspired by the recent advancements in pillar-based 3D object 
detectors~\cite{shi2022pillarnet,zhou2023fastpillars} due to their real-time speed and high performance. Therefore, we convert the sparse and irregular point clouds into dense and regular pillar representations. Specifically, pillar-based point cloud representation possesses following advantages: \textbf{(1) Pillar representation is dense and ordered}, facilitating seamless integration with advanced 2D image-based techniques without much modification. \textbf{(2) The compact nature of the pillar representation reduces computational overhead} while maintaining a desirable trade-off between performance and speed. \textbf{(3) Pillar representation is deployment-friendly}, making it highly suitable for resource-limited devices like mobile robots or drones. Particularly, pillar 
representation naturally aligns with the high demanding real-time requirements of 3D 
trackers, making it well-suited for tracking tasks.

Currently, there are few works~\cite{smat,pttr_journal} are dedicated to pillar-based 3D SOT tasks. SMAT~\cite{smat} addresses this gap by transforming sparse 3D point clouds into dense pillars and then utilizing a transformer-based encoder for multi-scale feature extraction and interaction, yielding promising results. However, SMAT utilizes a naive pillar encoding module derived from PointPillars~\cite{pointpillars}, which retains the substantial numerical disparities between point cloud channels. This approach doesn't fully leverage the intrinsic structure of the point cloud data and limit the representation robustness of pillar. Furthermore, its backbone directly adheres to the design principles of visual transformers~\cite{pvt,pvtV2} that are tailored for RGB images, which is not optimal for the point cloud modality. PTTR++~\cite{pttr_journal} introduces point-wise and pillar-wise view fusion to further enhance the performance of PTTR~\cite{zhou2022pttr}. Nonetheless, PTTR++ still incorporates a re-sampling operation in its process, introducing additional computational overhead and posing challenges for practical deployment.

In this paper, we propose PillarTrack, a pillar-based 3D single-object tracking framework, with the objective of augmenting tracking performance and accelerating inference speed, thereby facilitating deployment. Firstly, to reduce information loss caused by re-sampling operation, we propose a Pyramid Encoding Pillar Feature Encoder (PE-PFE) design to transform sparse point clouds into dense and robust pillar representations. Secondly, we argue the design of existing Transformer-based backbone from the perspective of modality differences and propose an modality-aware Transformer-based backbone specifically tailored for point cloud modalities, achieving higher performance with fewer computation resource. Finally, we construct our PillarTrack based on above designs. Extensive experiments on the KITTI and nuScenes dataset demonstrate the superiority of our method. As shown in Figure \ref{performance}, PillarTrack achieves comparable  performance on the KITTI and nuScenes dataset, striking a better balance between speed and accuracy, and catering to diverse practical needs. In conclusion, the main contributions of this paper are as follows:

$\bullet$ \textbf{PE-PFE}: A Pyramid-Encoded Pillar Feature Encoder (PE-PFE) is designed to encode the point coordinate of each pillar with pyramid-type representation. This approach not only enhances performance without incurring additional computational overhead but also improves robustness against rotation, translation, and scale variations.

$\bullet$ \textbf{Modality-aware Transformer-based Backbone}: A tailored for point cloud modalities backbone design, aiming to enhance pillar representation. It involves an adjustment of computational resources allocated to the front-end of the backbone, enabling the capture of more inherent semantic details from the input point cloud.

$\bullet$ \textbf{Comparable and Open-source}: Extensive experiments on KITTI and nuScenes dataset shown the comparable performance, significantly improving upon the baseline approaches. Besides, we open source our code to the community in \url{https://github.com/StiphyJay/PillarTrack}.

\vspace{-1mm}
\section{Related Work}
\vspace{-1mm}

\noindent\textbf{Single Object Tracking on Point Clouds.} Leveraging the advantages of LiDAR, which is less sensitive to illumination changes and captures accurate distance information, numerous works\cite{sc3d, cui2019point, fang20203d, p2b, bat, mmtrack, v2b} in LiDAR-based 3D Single Object Tracking (3D SOT) have merged. SC3D\cite{sc3d} utilizes a Kalman filter for proposals generation and selects the most suitable candidate via a Siamese network. 3D-SiamRPN \cite{fang20203d} combines 3D Siamese network with 3D RPN network for tracking. P2B\cite{p2b} introduces target-specific features and employs VoteNet\cite{votenet} to estimate the target center. Based on this, BAT\cite{bat} incorporates box-aware information to enhance similarity feature. M2-Track \cite{mmtrack,zheng2023effective} adopts a motion-centric paradigm that yields impressive results. Inspired by the success of transformers\cite{vaswani2017attention}, several transformer-based methods \cite{ptt, ptt-journal,lttr, zhou2022pttr, guo2022cmt, hui2022stnet, glt, towards} have merged. PTT \cite{ptt,ptt-journal} proposed Point-Track-Transformer module to assign weights to crucial point cloud features. PTTR\cite{zhou2022pttr} employs an attention-based feature matching mechanism to integrate target clues in similarity. STNet\cite{hui2022stnet} develops an iterative coarse-to-fine correlation network for robust correlation learning. GLT-T\cite{glt} proposes a global-local Transformer voting scheme to generate higher quality 3D proposals. Recently, SMAT\cite{smat} encodes global similarity based on attention to alleviate the point sparsity. However, their backbone still partially inherits the design principles from ResNet\cite{resnet} tailored for the image modality, which is sub-optimal for point clouds. Therefore, we argue their design and propose a modality-aware transformer-based backbone tailored for point cloud modality.

\begin{figure*}[!htbp]
\centering
\includegraphics[width=0.9\textwidth]{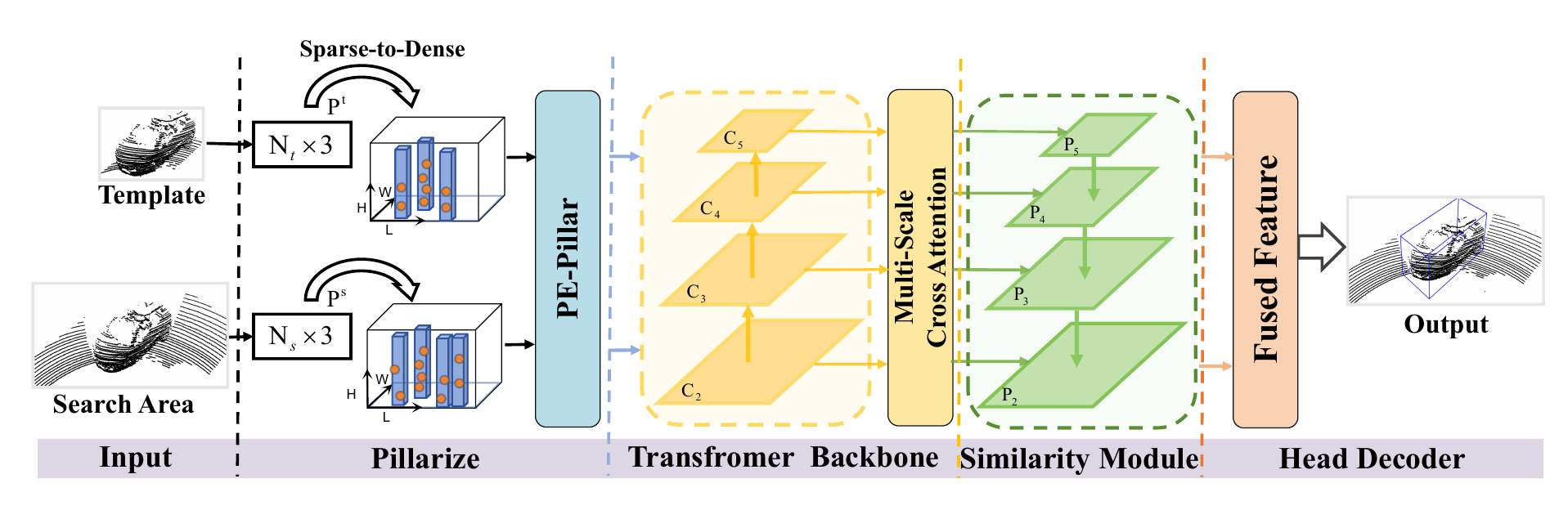}\vspace{-5mm}
\caption{The architecture of our PillarTrack network. Given the template and search area, we first use PE-PFE to extract multi-scale features respectively. Then, the backbone extract features at different feature scale and fuses the multi-scale similarity feature. Finally, we apply the detection head on the feature fusion map to localize the target.}
\label{fig:num_blocks}
\vspace{-6mm}
\end{figure*}

\noindent\textbf{Vision Transformer.} The Transformer architecture~\cite{vaswani2017attention}, originally proposed for natural language processing (NLP), has demonstrated significant success in image recognition due to its global modeling capabilities. The Vision Transformer (ViT)\cite{vit} classifies images by dividing them into patches, while Swin Transformer\cite{swin} introduces a shifted window attention mechanism, achieving excellent performance across various visual tasks. PVT~\cite{pvt} and PVTv2~\cite{pvtV2} employ a progressive shrinking pyramid to construct a visual Transformer backbone. In 3D vision, Point Transformer~\cite{zhao2021point} features a specialized Point Transformer layer for processing unordered point clouds. These methods have shown outstanding performance in point cloud shape classification and part segmentation. In contrast, our approach explores the potential of the Transformer's global modeling on pillar features in the object tracking task.

\vspace{-2mm}
\section{Preliminary}
\noindent\textbf{Motivation.} Our research endeavors to investigate a real-time, high-performance, pillar-based approach for 3D single-object tracking. To achieve this, we leverage the compact pillar representation and design a strong Transformer-based backbone that enhances pillar feature representation with reduced computational resources from the perspective of inherent modality differences. Regarding the feature integration and similarity computation module, we follow the original design of baseline\cite{smat}, which has shown remarkable efficacy. We hope that our designs can offer a novel perspective in 3D SOT, encouraging the community to revisit existing transformer-based methods.

\noindent\textbf{Task Definition.} In the task of the 3D Single Object Tracking (3D SOT), we initially define target point cloud (i.e., template) $\mathbf{P_t}=\{\mathbf{p}_{i}^{t}=[x_i, y_i, z_i, r_i]^T \in \mathbb{R}^{N \times 4}\}$, where $x_i, y_i, z_i$ denote the coordinate values of each point along the axes X, Y, Z, respectively, and $r_i$ represents the laser reflection intensity. Our goal is to locate the the 3D bounding box (3D BBox) of this target in the search area $\mathbf{P_t}=\{\mathbf{p}_{si}=[x_j, y_j, z_j, r_j]^T \in \mathbb{R}^{M \times 4}\}$ frame by frame. Here, $N$ and $M$ is the number of points in target and search area. The 3D BBox is represented as $\mathbf{B}=\{\mathbf{b}=[x, y, z, h, w, l, {\theta}]^T \in \mathbb{R}^{1 \times 7}\}$, The coordinates $x, y, z$ indicate the object's center, while $h, w, l$ denote its size, ${\theta}_j$ is the object's heading angle. Since the target's 3D BBox is known in the first frame, we need to regress the target center and heading angle in subsequent frames. By applying the displacement and heading angle from the previous frame's 3D BBox, we can locate the target in the current frame.

\noindent\textbf{Baseline Overview.} As SMAT is the first attempt at a pillar-based 3D Siamese tracker and has shown promising results, we choose SMAT as our baseline and briefly review its structure. The core components of SMAT include: \textbf{1)} a naive pillar feature encoding module derived from Pointpillars, which transforms raw point cloud data into a pseudo-image; \textbf{2)} a Transformer-based backbone network that computes similarity through a multi-head attention mechanism for the multi-scale features from both the template and search branches, capturing local and global information within the point cloud; \textbf{3)} a neck module that integrates features from different scales and fuses multi-scale similarity features, enhancing the model's perception of multi-scale information; \textbf{4)} a decoder that predicts the position and orientation of the target object. These designs enable SMAT to effectively handle sparse point cloud, offering an efficient solution for 3D SOT.


\vspace{-2mm}
\section{Method}

\subsection{Pyramid-Encoded Pillar Feature Encoding} 

\begin{figure}[!htbp]
\vspace{-3mm}
\centering
\includegraphics[width=0.9\linewidth]{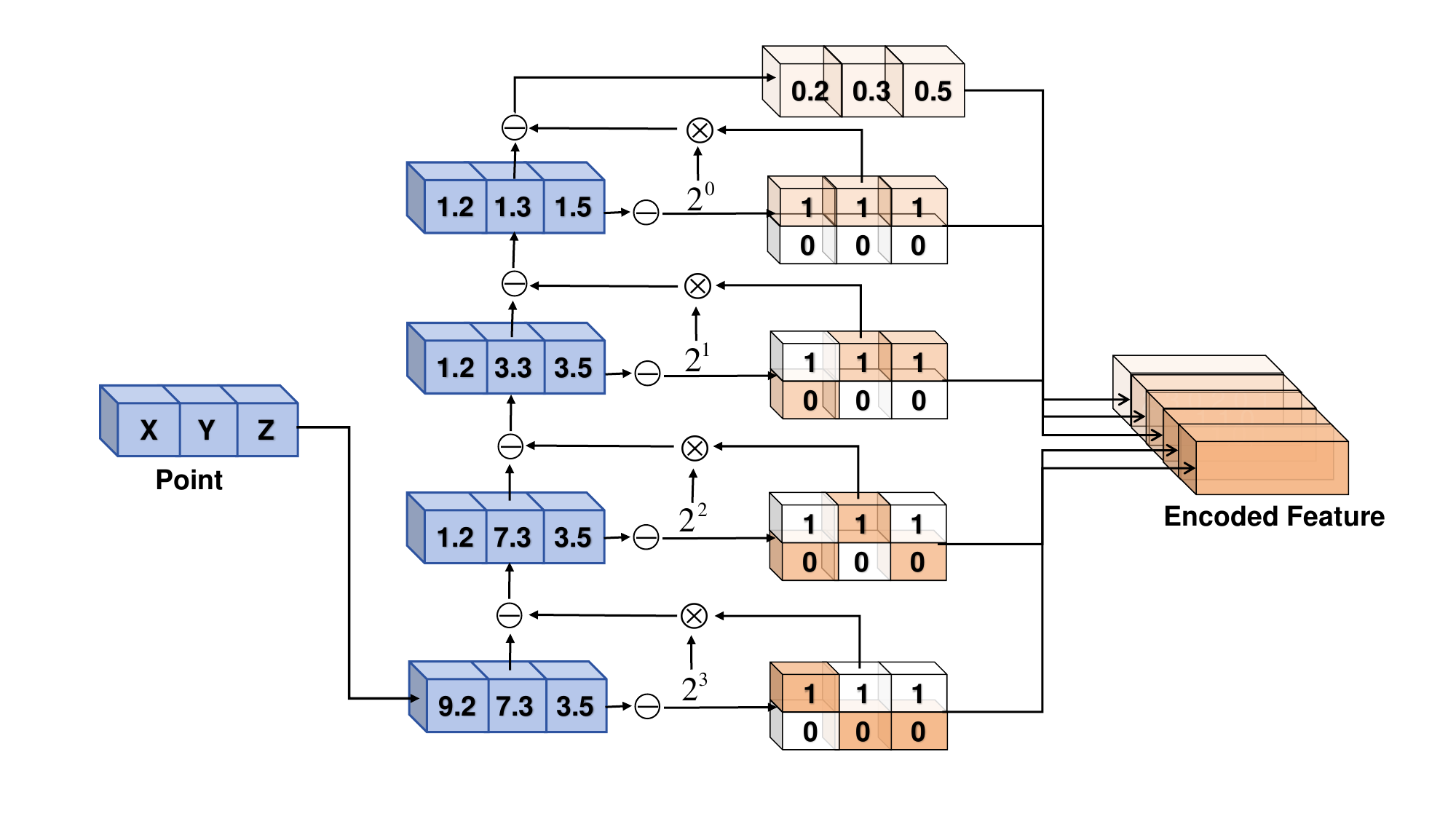}
\vspace{-3mm}
\caption{The illustration of PE-PFE design. We encode the point coordinate of the input point cloud in a pyramid-like type. This Pyramid-like encoding design allows the network to optimize effectively without the input information loss.}
\label{fig:PE-PFE}
\vspace{-2mm}
\end{figure}

\begin{figure*}[h]
\centering
\includegraphics[width=1\textwidth]{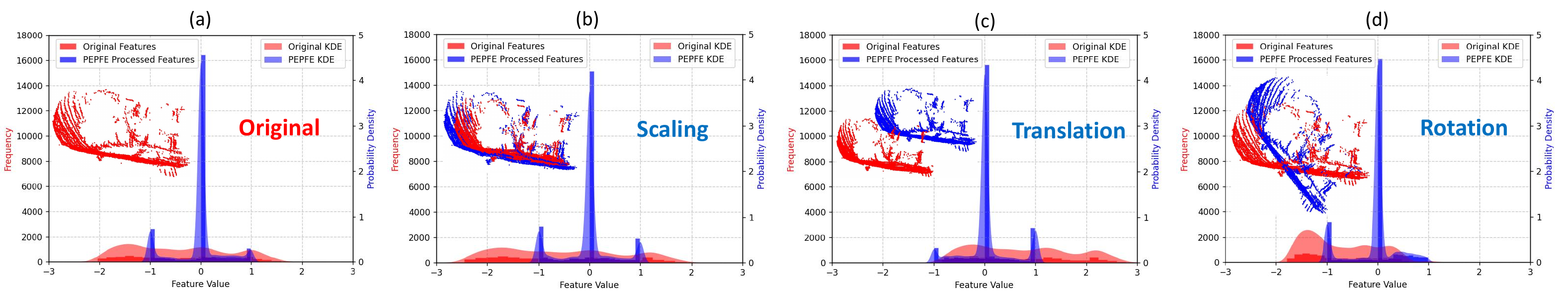}\vspace{-2mm}
\caption{Illustration of the properties of PE-PFE. The \textcolor{red}{red} indicates the feature distribution after original PFE encoding, while the \textcolor{blue}{blue} represents the feature distribution after PE-PFE encoding. (a) Original point cloud. (b) Point cloud with 1.2X scale. (c) Point cloud with 1.2m translation. (d) Point cloud with 45$\degree$ rotation.}
\label{fig:pepfe-robust}
\vspace{-4mm}
\end{figure*}

\begin{figure*}[h]
\centering
\includegraphics[width=1\textwidth]{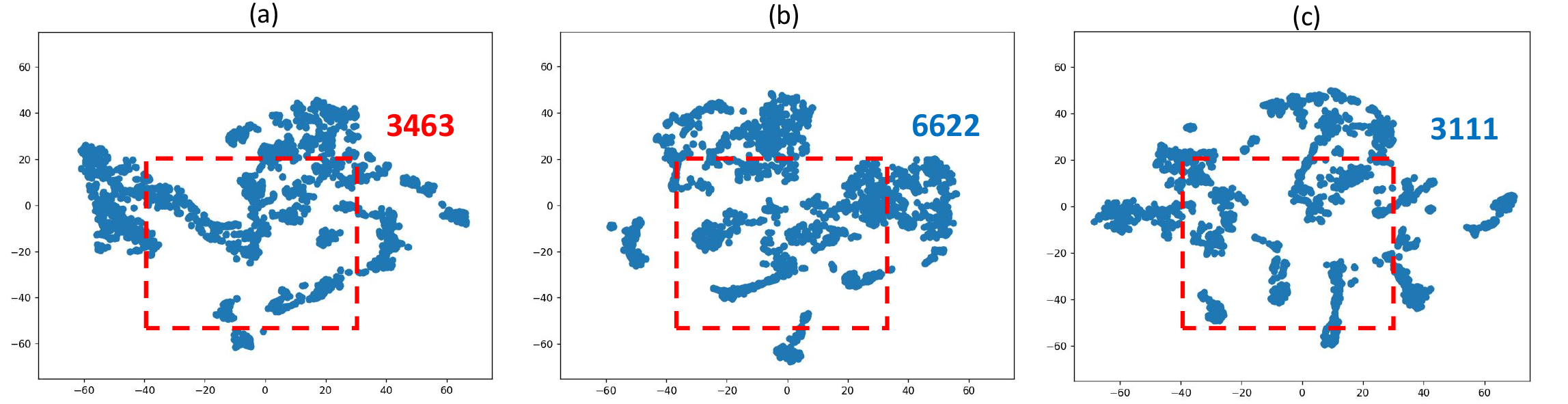}\vspace{-2mm}
\caption{The output features visualization of forth stage in t-SNE.(a) Visualization with the image blocks setting. (b) Visualization matching the total number of the original blocks. (c) Visualization after removing redundancies from the number of blocks.}
\label{fig:TSNE}
\vspace{-5mm}
\end{figure*}

In pillar-based 3D detection methods\cite{pointpillars,shi2022pillarnet,zhou2023fastpillars}, Pillar Feature Encoding (PFE) plays a crucial role, directly impacting the representation learning of BEV feature and final detection accuracy. Its significance also extends to 3D single object tracking tasks. SMAT utilizes the PFE from PointPillars~\cite{pointpillars} to extract pillar-wise feature. Here, taking PFE module as an example, we illustrate the details of their input encoding. The points in each pillar are initially concatenated along the channel dimension to form $\hat{p_i}=\{[x_i, y_i, z_i, r_i,  {x}_{i}^{m}, {y}_{i}^{m}, {z}_{i}^{m}, {x}_{i}^{c},{y}_{i}^{c}, {z}_{i}^{c}] \in {\mathbb{R}^{N_v \times 10}} \}$, where $[x_i, y_i, z_i]$ is the original point coordinates in the ego frame, $N_v$ is the number of points within the pillar, $[{x}_{i}^{m}, {y}_{i}^{m}, {z}_{i}^{m}]$ denotes the offset of $p_i$ from the mean of the xyz coordinates of all points within the current pillar, and $[{x}_{i}^{c},{y}_{i}^{c}, {z}_{i}^{c}]$ is the offset of $p_i$ from the current pillar center. However, there exist large numerical differences between the feature channels in PFE. For instance, the numerical range of $[{x}_{i}^{m}, {y}_{i}^{m}, {z}_{i}^{m}]$ and $[{x}_{i}^{c},{y}_{i}^{c}, {z}_{i}^{c}]$ is $[-0.1, 0.1]$, since they are relative offset within a pillar. Such huge disparities in input values pose significant challenges for network optimization~\cite{santurkar2018does} and can make model sensitive to disturbance, such as scaling, translation or rotation. To this end, we propose the Pyramid-Encoding Pillar Feature Encoder (PE-PFE), which is designed to minimize value discrepancies between channels, thereby facilitating improved optimization of numerically stable inputs. Besides, PE-PFE ensures that input point clouds maintain robustness against scale, translation, rotation.

\vspace{-6.3mm}
In detail, as shown in Figure \ref{fig:PE-PFE}, we encode the channel values of the input point cloud in a pyramid-type manner, and finally retain a residual value that cannot be evenly divided. The encoding values from each layer and the residual value are then cascaded together as point-wise features, which are then processed by the PFE module. Similar to the natural language processing (NLP) tokenization, the PE-PFE serves as a feature embedding that converts point cloud data into an information-dense representation. The embedding of PE-PFE is characterized by threes properties: scale-robustness, translation-robustness, and rotation-robustness. Scale robustness aids in the feature extraction of geometrical information for vehicles with identical shapes but varying sizes, while remaining insensitive to scale variations. Translation-robustness ensures that even with positional perturbations in the input point cloud, the distribution of feature encoding remains stable, facilitating reliable representation. Rotation-robustness assists in extracting point cloud with different orientation angle, enhancing the expression of semantics. Collectively, these properties contribute to the robustness and effectiveness of PE-PFE in feature extraction and geometric understanding within the 3D SOT tasks. For instance, as show in Figure \ref{fig:pepfe-robust}, we apply different operations of 1.2X scale, 1.2m translation, and 45$\degree$ rotation to the raw point cloud of a vehicle. The input feature distribution of original PFE exhibits significant discrepancies; however, after PE-PFE, the feature distributions remained more stable. We remain the exploration of PE-PFE on other Lidar-based tasks in the future work. For the effectiveness of PE-PFE, refer to Table \ref{table:pepfe ablation experiment}. 
\subsection{Modality-aware Transformer-based Backbone} 

Recent studies~\cite{tian2022fully, zhou2023fastpillars, LKUnet} have pointed out that the backbone designs commonly employed in image domain~\cite{resnet, swin} are sub-optimal for point cloud or depth images. This is due to the significant modality differences between RGB images and LiDAR point clouds, where LiDAR inherently capture geometric semantics. Particularly, backbone design in image domain tends to utilize more computation burden in later stages to extract high-level semantic features for perception tasks. For instance, ResNet-50\cite{resnet} takes 6 residual blocks in the 3rd stage while using 3 blocks in 1st stage for feature extraction. Similarly, Swin-T~\cite{swin} adopts 9 blocks in the 3rd stage while only 1 block is used in the 1st stage. However, we argue that these design choices for distributing computations across stages were largely empirical and not be well-suited for LiDAR point clouds. Unlike RGB images, LiDAR can capture spatial distances and shapes of objects. This means that rich and accurate geometric information of objects is already explicitly captured in raw point cloud coordinates. Therefore, we assume that instead of allocating excessive computational resources to model object geometries in later stages like ResNet/Swin-T in RGB images, we propose reallocating more computational resources to the early stages to effectively incorporate the geometric information carried inherenty by point clouds.

To validate the aforementioned hypothesis for a transformer-based backbone, we performed t-SNE\footnote{t-SNE (t-distributed Stochastic Neighbor Embedding) is a machine learning algorithm designed for the visualization of high-dimensional data. It maps high-dimensional data to a lower-dimensional space while preserving relative distances and distribution structures, thereby facilitating exploratory data analysis and providing insights into the internal structure of the data.} visualization~\cite{van2008tsne} on the features processed at different stages. The effectiveness of feature extraction can be evaluated by observing cluster separation, shape, and compactness in the t-SNE visualization. Clearly separated, regular, and compact clusters suggest successful feature extraction. Initially, we visualized the features under the original blocks number setting using t-SNE, as indicated in the red box of Figure \ref{fig:TSNE} (a). The visualization revealed significant overlap between clusters, irregular shapes, and lack of compactness, indicating that the features were not adequately processed under the original configuration. Therefore, keeping the total number of blocks constant, we allocated more computational resources to the earlier stages of the network. As shown in the red box of Figure \ref{fig:TSNE} (b), this adjustment led to less overlap between clusters compared to the original setting, demonstrating some improvement and the learned diverse representation. However, the clusters still lacked compactness, leading us to hypothesize that there was redundant computation. To test this hypothesis, we conducted a comparative experiment with a "3111" block configuration, as illustrated in the red box of Figure \ref{fig:TSNE} (c). This setup resulted in clearly separated clusters, many of which exhibited an elliptical shape, suggesting that our hypothesis was preliminarily confirmed. In summary, the adjustments made to the allocation of computational resources across the stages of the transformer-based backbone have significantly improved the quality of feature extraction, as evidenced by the enhanced cluster separation and regularity observed in the t-SNE visualizations.

\begin{table*}[t]
\caption{Performance comparison on the KITTI dataset. L means Lidar-only methods. M and S are Motion-based and Similarity-based paradigms. The best two results are highlighted in \textcolor{red}{red}, \textcolor{blue}{blue}. * indicates results produced based on the official
implementation.}
\centering
\resizebox{\linewidth}{!}{
\begin{tabular}{c|c|c|c|cc|cc|cc|cc|cc|cc}
		\hline
		\multirow{2}{*}{Method} & \multirow{2}{*}{Publication} & \multirow{2}{*}{Paradigm} & \multirow{2}{*}{View}  & \multicolumn{2}{c|}{Car-6424} & \multicolumn{2}{c|}{Pedestrian-6088} & \multicolumn{2}{c|}{Van-1248} & \multicolumn{2}{c|}{Cyclist-308} & \multicolumn{2}{c|}{Mean-by class} & \multicolumn{2}{c}{Mean-by frame}  \\
  & & & & Success    & Precision & Success    & Precision & Success    & Precision & Success    & Precision & Success    & Precision & Success    & Precision\\\hline 
	$M^2$-Track*  ~\cite{mmtrack} & CVPR 2022 & L+M &Point & 64.8 & 79.2 & \textcolor{blue}{57.6} &\textcolor{blue}{83.6} & 50.4 & 64.8 & \textcolor{blue}{74.9} & 93.3 &61.9 & 80.2 &60.6  &\textcolor{blue}{80.1}  \\ \hline
		SC3D\cite{sc3d} & CVPR 2019 &\multirow{15}{*}{L+S}  &Point & 41.3       & 57.9        & 18.2       & 37.8        & 40.4       & 47.0        & 41.5       & 70.4 & 35.4 & 53.3 & 31.2 & 48.5 \\
  P2B\cite{p2b} & CVPR 2020 &  &Point & 56.2& 72.8 & 28.7 & 49.6 & 40.8  & 48.4  & 32.1 & 44.7  &39.5 & 53.9 & 42.4 & 60.0 \\
  3D-SiamRPN\cite{fang20203d} & IEEE Sensors J &   &Point & 58.2 & 76.2 & 35.2 & 56.2 & 45.7 & 52.9 & 36.2 & 49.0  &43.8 & 58.6 &46.6 & 64.9 \\
  MLVSNet\cite{MLVSNet} & ICCV2021 &   &Point &56.0 & 74.0 & 34.1 & 61.1 & 52.0 & 61.4 & 34.3 & 44.5 &44.1 & 60.3 & 45.7 & 66.7 \\
  BAT\cite{bat}  & ICCV 2021 &   &Point  & 60.5       & 77.7        & 42.1       & 70.1        & 52.4       & 67.0& 33.7       & 45.4  &47.2 & 65.1 & 51.2 & 72.8\\
  PTT\cite{ptt}  & IROS 2021 &   &Point & 67.8       & 81.8        & 44.9       & 72.0        & 43.6       & 52.5        & 37.2       & 47.3 &48.4 & 63.4& 55.1 & 74.2\\
  LTTR\cite{lttr}  & BMVC 2021 &   &Voxel & 65.0       & 77.1        & 33.2       & 56.8        & 35.8       & 45.6        & 66.2      & 89.9 &50.0 & 67.4& 48.7 & 65.8\\
  V2B\cite{v2b}  &NeurIPS 2021  &  &Point+Voxel & 70.5       & 81.3        & 48.3       & 73.5        & 50.1       & 58.0        & 40.8       & 49.7 &52.4 & 65.6 &58.4  &75.2 \\
  PTTR\cite{zhou2022pttr}  & CVPR 2022 &   &Point & 65.2       & 77.4        & 50.9       & 81.6        & 52.5& 61.8        & 65.1       & 90.5 &58.4 &77.8 & 57.9 & 78.1\\
  CMT\cite{guo2022cmt}  &  ECCV 2022&   &Point & 70.5       & 81.9        & 49.1       & 75.5        & 54.1& 64.1        & 55.1       & 82.4 &57.2 &76.0 &59.4  &77.6 \\
  STNet\cite{hui2022stnet}  &  ECCV 2022&   &Point &\textcolor{blue}{ 72.1}&\textcolor{blue}{84.0}& 49.9& 77.2& \textcolor{blue}{58.0}& \textcolor{red}{70.6}& 73.5 & \textcolor{blue}{93.7} &\textcolor{blue}{63.4}&\textcolor{blue}{81.4} &\textcolor{blue}{61.3}  &\textcolor{blue}{80.1} \\
        C2FT\cite{c2ft}  &  RAL 2022&  &Point & 67.0& 80.4& 48.6& 75.6& 53.4& 66.1& 38.0& 48.7 &51.6 &67.7 &57.2 &76.4 \\
        MLSET\cite{mlset}  &  RAL 2023&   &Point & 69.7& 81.0& 50.7& 80.0& 55.2& 64.8& 41.0& 49.7 &54.2 &68.9 &59.6  &78.4  \\
        \hline
        \rowcolor{gray!30}
        Baseline* \cite{smat} & RAL 2022 &   &Pillar & 71.9& 82.4& 52.1& 81.5& 41.4& 53.2& 61.2& 87.3 &56.7 &76.1 & 60.4 & 79.5 \\
        \rowcolor{gray!55}
        \textbf{PillarTrack}*   & Ours &   &Pillar &\textcolor{red}{74.2}   &\textcolor{red}{85.1} &\textcolor{red}{59.7}  &\textcolor{red}{84.7}  &\textcolor{red}{61.0}  &\textcolor{blue}{69.2} &\textcolor{red}{78.0}  &\textcolor{red}{95.0}  &\textcolor{red}{68.2}  &\textcolor{red}{83.5}  &\textcolor{red}{66.8}   &\textcolor{red}{83.7} \\
		\hline
  Improvement &  & & & \textcolor{green}{+2.3} & \textcolor{green}{+2.7} & \textcolor{green}{+7.6} &\textcolor{green}{+3.2} & \textcolor{green}{+19.6}&\textcolor{green} {+16.0}& \textcolor{green}{+16.8}&\textcolor{green}{+7.7} &\textcolor{green}{+11.4} & \textcolor{green}{+7.2}  &\textcolor{green} {+6.4} &\textcolor{green} {+4.2}\\
  \hline
    \end{tabular}
    }
\label{tab:kitti}
\vspace{-5mm}
\end{table*}

To further substantiate our hypothesis from the results, we systematically study computation allocation by adjusting the stage compute ratio in the transformer-based backbone of SMAT~\cite{smat}. It is worth noting that the setting of SMAT has the same depth number as ResNet-50, with a stage compute ratio of $[3, 4, 6, 3]$. As shown in Table \ref{fig:backbone}, we used the SMAT backbone as the baseline and varied the number of depth blocks from 0 to 4 with a stride of 1 in each stage. From the results, we can find that the performance is quite sensitive to the capacity of front stage, while being less sensitive to the later stages (2, 3, 4). This confirmed that allocating more computational resources to the early stages is beneficial for enabling the network to better capture the geometric information carried by the raw point clouds. Based on this observation, we set the stage compute ratio of the four stages to $[3, 1, 1, 1]$. It is worth noting that with this depth blocks setting, our modality-aware transformer backbone network achieved promising performance improvements in terms of 3.2/\%  in Success and 2.5/\% in Precision metric, while significantly reducing the computational cost by half the GFLOPs compared to the baseline method (from 0.37 to 0.18). It is noteworthy that an increase in computational load at the front end leads to a saturated performance. We surmise this is due to the saturation of the network's parameters, where the excessive employment of parameters augments the learning burden on the network. This phenomenon naturally substantiates the existence of a performance bottleneck in the principle of computational resource allocation, illustrating that performance does not escalate linearly with additional computational resources.

\begin{table*}[t]
\vspace{0.2in}
\caption{Performance comparison on the NuScenes dataset. The best two results are highlighted in \textcolor{red}{red}, \textcolor{blue}{blue}.}
\renewcommand\tabcolsep{3pt}
\scriptsize
\centering
\begin{tabular}{c|c|cc|cc|cc|cc|cc|cc|cc}
\toprule[.05cm]
\multirow{2}{*}{Method} & \multirow{2}{*}{Paradigm} & \multicolumn{2}{c|}{Car-64159} & \multicolumn{2}{c|}{Pedestrian-33227} & \multicolumn{2}{c|}{Truck-13587} & \multicolumn{2}{c|}{Trailer-3352} & \multicolumn{2}{c|}{Bus-2953} &\multicolumn{2}{c|}{Mean by class} & \multicolumn{2}{c}{Mean by frame} \\
&                           & Success    & Precision   & Success    & Precision   & Success     & Precision    & Success      & Precision    & Success    & Precision & Success    & Precision & Success    & Precision  \\ \hline \hline
$M^2$-Tracker~\cite{mmtrack}  & Motion & \textcolor{red}{55.85} &\textcolor{red}{65.09}   &       32.10      &  \textcolor{blue}{60.92}  &\textcolor{red}{57.36} &\textcolor{red}{59.54}      &\textcolor{blue}{  57.61}  &\textcolor{red}{58.26}   &  \textcolor{red}{51.39}  &\textcolor{red}{51.44} &\textcolor{red}{50.86}&\textcolor{red}{59.05} &\textcolor{red}{49.23}&\textcolor{red}{62.73} \\ \hline
SC3D~\cite{sc3d} & \multirow{8}{*}{Similarity}    &   22.31         &    21.93  &11.29   &12.65    & 30.67  &27.73    &35.28  & 28.12  & 29.35  &24.08  & 25.78&22.90 &20.70&20.20            \\
P2B~\cite{p2b} &   & 38.81  & 43.18  &  28.39 & 52.24  & 42.95 & 41.59&  48.96     &  40.05    &  32.95 & 27.41 &38.41 &40.90 &36.48&45.08  \\
PTT~\cite{ptt-journal} &   & 41.22  & 45.26  &  19.33 & 32.03  & 50.23 & 48.56&  51.70     &  46.50    &  39.40 & 36.70 &40.38 &41.81 &36.33&41.72  \\
BAT~\cite{bat} &   &40.73 & 43.29 & 28.83 & 53.32   &45.34&   42.58   &52.59   &44.89  &35.44   & 28.01 &40.59 &42.42  &38.10&45.71  \\ 
GLT-T~\cite{glt} &   & 48.52  & 54.29  &  31.74 & 56.49  & 52.74 & 51.43&  57.60     &  52.01    &  44.55 & 40.69 &47.03 &50.98 &44.42&54.33  \\
PTTR~\cite{zhou2022pttr} &   &\textcolor{blue}{ 51.89}  &\textcolor{blue}{ 58.61}  &  29.90 & 45.09  & 45.30 & 44.74&  45.87     &  38.36    &  43.14 & 37.74 &43.22 &44.91  &44.50&52.07  \\
\hline
\rowcolor{gray!25}
Baseline~\cite{smat} &   & 43.51 & 49.04 & \textcolor{blue}{32.27}  &60.28 & 44.78& 44.69&37.45 & 34.10  &39.42 & 34.32 &39.49 &44.49 &40.20&50.92            \\ 
\rowcolor{gray!50}
\textbf{PillarTrack} & Ours & 47.12 & 57.72 &\textcolor{red}{34.18} &\textcolor{red}{64.93}& \textcolor{blue}{54.82}& \textcolor{blue}{54.41}&\textcolor{red}{57.70} &\textcolor{blue}{54.63} &\textcolor{blue}{44.68} &\textcolor{blue}{40.73} & \textcolor{blue}{47.70}& \textcolor{blue}{54.48} &\textcolor{blue}{44.59} &\textcolor{blue}{58.86}\\ \hline
Improvement &  & \textcolor{green}{+3.61} & \textcolor{green}{+8.68} & \textcolor{green}{+1.91} &\textcolor{green}{+4.65} & \textcolor{green}{+10.04}&\textcolor{green} {+9.72}& \textcolor{green}{+17.25}&\textcolor{green}{+20.53} &\textcolor{green}{+5.26} & \textcolor{green}{+6.41}  &\textcolor{green} {+8.21} &\textcolor{green} {+9.99}  &\textcolor{green} {+4.39} &\textcolor{green} {+7.94}\\
\toprule[.05cm]
\end{tabular}
\label{tab:nuscenes}
\vspace{-3mm}
\end{table*}

\textbf{Backbone Designs Differences with Previous Study.} Our results differ significantly from previous studies~\cite{tian2022fully, zhou2023fastpillars}, which were focused on LiDAR-based object detection tasks using CNN architecture with local receptive fields. In contrast, we focus on 3D SOT tasks within local search area and design a transformer architecture with global receptive fields. Therefore, by allocating more computational resources to the 1st stage, our modality-aware transformer-based backbone is able to capture the enough global-aware geometric semantics of the input point cloud. \textbf{This is very different with CNN-based backbone, which require more computation overhead allocated to both the 1st, 2nd and 3rd stages.} 
\vspace{-2mm}
\section{Experiments}
\noindent\textbf{Dataset and Evaluation metrics.} Following previous works\cite{sc3d,p2b,bat,pttr_journal,mmtrack,smat}. we evaluate our method on KITTI \cite{kitti} and nuScenes \cite{nuscenes} datasets. For KITTI datasets, we follow the same setting in baseline model\cite{smat} to divide the sequences into training, validation, and testing splits, where 0-16 for training, 17-18 for validation and 19-20 for testing. Notably, the nuScenes\cite{nuscenes} is a more challenging dataset due to less channel (32-line) in LiDAR and more complex scenes, containing 1000 scenes, which are allocated into 700/150/150 scenes for training, validation, and testing, respectively. For metrics, we follow the baseline and use One-Pass Evaluation (OPE) \cite{kristan2016novel} to measure Success and Precision. The Success measure the 3D IoU between the predicted box and the ground truth box, the Precision measures the AUC of the distance error between the center of the predicted and ground truth box.

\noindent \textbf{Implementation details.}
We follow our baseline model\cite{smat} and employ PVTv2-b2 \cite{pvtV2} as backbone. The specific settings include a head number of [1, 2, 5, 8], a depth configuration of [3, 1, 1, 1], the expansion ratio for the feed-forward layer set to [8, 8, 4, 8] and a channel number configuration of [64, 128, 320, 512]. In KITTI, we set [0.05, 0.05, 4] meters as pillar size and a search area of [-3.2, -3.2, -3, 3.2, 3.2, 1] meters for the car category. In nuScenes, we set [0.075, 0.075, 8] meters as pillar size and a search area of [-3.2, -3.2, -5, 3.2, 3.2, 3] meters for the car category. To fair comparison, other training setting is the same as our baseline model, more details refer to SMAT~\cite{smat}. 
\vspace{-2mm}
\subsection{Comparison with State-of-the-art Trackers}

\noindent\textbf{Quantitative results on KITTI.} As shown in Table \ref{tab:kitti}, the proposed PillarTrack demonstrates significant performance improvements compared to the baseline method (SMAT \cite{smat}). Specifically, PillarTrack achieved a performance improvement of 11.4\% \/ 7.2\% and 6.4\% \/ 4.2\% in Success and Precision metrics in terms of mean by-class and mean-by frame setting, respectively. In the categories of the Car, Pedestrian, Van, and Cyclist, PillarTrack outperformed SMAT by 2.3\%, 7.6\%, 19.6\%, and 16.8\% in Success, respectively. Besides, PillarTrack also surpassed the motion-based method MMTrack~\cite{mmtrack} in all classes, achieving promising results on KITTI dataset. 

\noindent\textbf{Quantitative results on NuScenes.} The KITTI dataset has a limited number of samples, resulting in relatively modest performance gains, which may not fully reflect the capabilities of the model. Therefore, we also conduct experiments on the large-scale nuScenes~\cite{nuscenes} dataset. NuScenes contains many complex scenes with sparser point clouds, making it a more challenging benchmark. As shown in Table \ref{tab:nuscenes}, our PillarTrack achieves the second-best performance across the five categories on the nuScenes, demonstrating comparable performance among similarity-based methods. Notably, PillarTrack exhibits significant improvements over the baseline method \cite{smat} in each category, particularly with a 10.04\% and 9.72\% enhancement in Success and Precision metrics for the Truck category, and a 17.25\% and 20.52\% improvement for the Trailer category. However, our performance in the Car category is slightly lower than that of PTTR. We attribute this to PTTR's utilization of both point-based and pillar-based representations for encoding the input point cloud, which leads to superior performance. Nevertheless, their point-based configuration may limit the potential for further quantization acceleration in onboard devices. Notably, M2-Tracker \cite{mmtrack} adopts a motion-based paradigm for object tracking, which results in performance that surpasses similarity-based paradigms.

\begin{table}[h]
\vspace{-1mm}
\caption{Performance comparison of PE-PFE after noise disturbance. Baseline: original setting. Rotation: Add random rotation noise within -45° to 45°. Scale: Add random scaling from 0.8 to 1.2x for the input point clouds. Translation: Add random translation from -1.2m to 1.2m for the baseline.}
\resizebox{\linewidth}{!}{
\centering
\begin{tabular}{c|cc|cc|cc|cc}\toprule[.02cm]
\multirow{2}{*}{PFE Method}  & \multicolumn{2}{c|}{Baseline}  &  \multicolumn{2}{c|}{Scale}  & \multicolumn{2}{c|}{Translation} & \multicolumn{2}{c}{Rotation} \\ 
& Suc.    & Prec.   & Suc.    & Prec.   & Suc.    & Prec.   & Suc.    & Prec.\\
\toprule
PFE (SMAT\cite{smat}) & 68.9    & 79.6      & 70.9    & 81.5   & 68.5    & 78.5 & 69.6    & 79.9\\
PE-PFE (Ours) & 70.8    & 80.3      & 72.9    & 82.4   & 73.3    & 83.0 & 72.0    & 81.5\\
\toprule
\end{tabular}
}
\label{table:pepfe data augmentation}
\vspace{-2mm}
\end{table}

\begin{table}[h]
\centering
\vspace{-2mm}
\caption{Ablation of modality-aware transofrmer-based backbone. \textbf{Bold} denotes the best performance.}\label{table:template}
\resizebox{0.9\linewidth}{!}{
\begin{tabular}{c|c|c|c}\toprule[.02cm]
Blocks Number &GFLOPs & Success & Precision \\
\hline \hline
$\left[ 3, 4, 6, 3 \right]$ (SMAT\cite{smat}\footnotemark[4])   & 0.37  & 68.9  & 79.6 \\
\midrule
$\left[6, 6, 2, 2 \right]$  & 0.38  & 70.2  & 80.5 \\
\midrule
$\left[1, 1, 1, 1 \right]$  & 0.14  & 69.5  & 80.1 \\
\midrule
$\left[2, 1, 1, 1 \right]$   & 0.16  & 70.3  & 80.6   \\
\rowcolor{gray!55}
$\left[3, 1, 1, 1 \right]$    & \textbf{0.18} (\textbf{-0.19})  & \textbf{72.1} (\textbf{+3.2})  & \textbf{82.1} (\textbf{+2.5}) \\ 
$\left[4, 1, 1, 1 \right]$  & 0.21  & 68.7  & 78.9 \\ 
\midrule
$\left[1, 2, 1, 1 \right]$    & 0.16  & 70.1  & 80.7   \\
$\left[1, 3, 1, 1 \right]$   & 0.18  & 69.7  & 80.6   \\ 
$\left[1, 4, 1, 1 \right]$ & 0.20  & 69.4  & 79.4 \\
\midrule
$\left[1, 1, 2, 1 \right]$  & 0.16  & 68.4  & 78.8   \\
$\left[1, 1, 3, 1 \right]$  & 0.18  & 70.0  & 80.8   \\ 
$\left[1, 1, 4, 1 \right]$ & 0.20  & 71.0  & 81.8 \\
\midrule
$\left[1, 1, 1, 2 \right]$   & 0.15  & 68.0  & 78.3   \\
$\left[1, 1, 1, 3 \right]$  & 0.17  & 68.1  & 78.4   \\ 
$\left[1, 1, 1, 4 \right]$   & 0.18  & 69.3  & 80.0   \\ 
\bottomrule
\end{tabular}
}
\vspace{-2mm}
\label{fig:backbone}
\end{table}

\vspace{-3mm}
\subsection{Ablation Study}
\noindent\textbf{The robustness of Pe-PFE.} As shown in Tab.~\ref{table:pepfe data augmentation}, we evaluated the performance of our PE-PFE under different noise disturbance, with a particular focus on its robustness against scale, translation, and rotation variations. The results demonstrate that PE-PFE outperforms the baseline method PFE (SMAT\cite{smat}) across all conditions. These results indicate that PE-PFE effectively maintains performance stability in the presence of various noise disturbances, showcasing strong robustness.

\noindent\textbf{The Design of Modality-aware Transformer-based Backbone.} We conduct experiments by arranging the number of blocks per stages from 0 to 4 with a stride of 1, while setting the number of blocks in other stages to 1, as shown in Tab.~\ref{fig:backbone}. The results prove our previous assumption: for pillar-based LiDAR 3D SOT tasks, we should reallocate the capacity to the early stages to better integrate the geometric information carried by the raw points, instead of allocating the capacity in the later stages like image domain. Therefore, to achieve a better performance under the constraint of low computation resource, we set the stage compute ratio to (3, 1, 1, 1). 

\begin{table}[h]
\vspace{-3mm}
\caption{Ablation experiment with Blocks Number and PFE Method in the Car category in KITTI.}
\resizebox{\linewidth}{!}{
\centering
\begin{tabular}{c|c|c|c|c}\toprule[.02cm]
\multirow{1}{*}{Method}  & \multicolumn{1}{c|}{Blocks Number} & \multicolumn{1}{c|}{PFE Method} &  \multicolumn{1}{c|}{Success}  & \multicolumn{1}{c}{Precision} \\ 
\toprule
SMAT\cite{smat} &$\left[3, 4, 6, 3 \right]$  &PFE &68.9 &79.6 \\ \toprule
\multirow{3}{*}{Ours} &$\left[3, 4, 6, 3 \right]$  &PEPFE &70.8 &80.3 \\
 &$\left[3, 1, 1, 1 \right]$  &PFE &72.1 &82.1 \\
&$\left[3, 1, 1, 1 \right]$  &PEPFE &74.2 &85.1 \\
\toprule

\end{tabular}
}
\label{table:pepfe ablation experiment}
\vspace{-2mm}
\end{table}

\noindent\textbf{Overall Ablation.} Here, we conducted an ablation study to investigate the impact of the proposed PE-PFE and modality-aware backbone network design on the overall performance, as shown in Table \ref{table:pepfe ablation experiment}. Firstly, when employing the same block configuration ($\left[3, 4, 6, 3 \right]$), the PE-PFE method achieved a success rate of 70.8\% and a precision of 80.3\%, surpassing the baseline method SMAT. Secondly, by adjusting the block configuration to $\left[3, 1, 1, 1 \right]$, we observed that the success rate of the PE-PFE method further increased to 74.2\%, with precision rising to 85.1\%. This indicates that reducing the number of blocks, in conjunction with the PE-PFE strategy, effectively enhances the model's tracking capability.

\begin{figure*}[h] 
    \centering
    \includegraphics[width=1\linewidth]{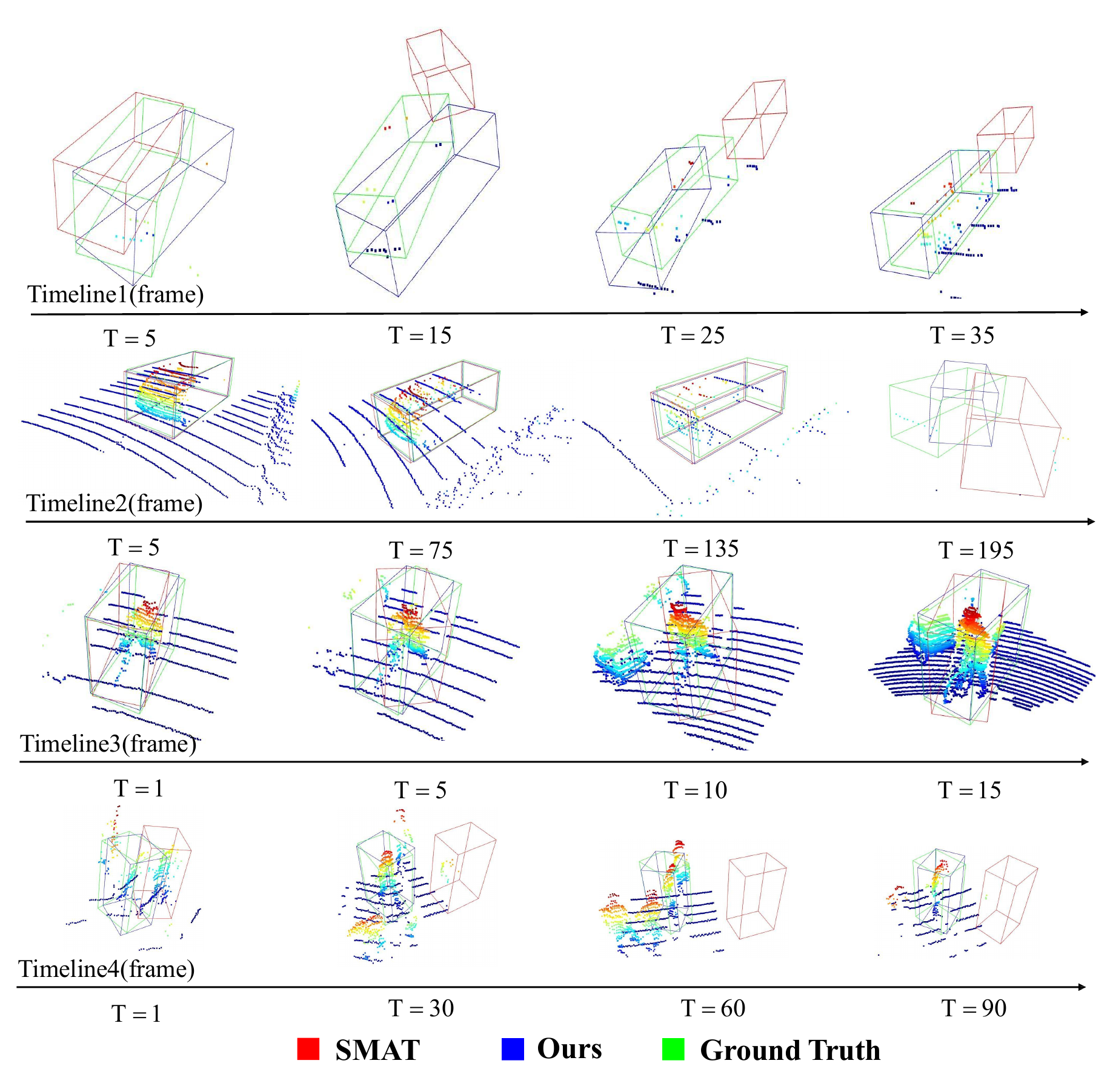}\vspace{-3mm}
    \caption{Visualization results comparison of our PillarTrack and baseline method (SMAT\cite{smat}) in KITTI dataset.}
    \label{fig:vis}
    \vspace{-5mm}
    \label{vis}
\end{figure*}

\vspace{-4mm}
\subsection{Visualization results.}
To comprehensively evaluate the performance of our method, particularly its more representative capability of point clouds, we conduct visualization on the different category, encompassing both sparse and dense scenes. Through comparative experiments on the initial frame and subsequent sequences, our method demonstrated superior tracking accuracy in both sparse and dense scenes compared to the baseline\cite{smat} method, highlighting ours enhanced adaptability and robustness. In the Cyclist category, our method was able to consistently track the target, whereas SMAT lost the target. For the Pedestrian category, our method maintained stable tracking from the outset, while SMAT failed to identify the targets. The visualization results in Figure \ref{fig:vis} substantiate the effectiveness and superior performance of our method across various scenarios.

\vspace{-3mm}
\subsection{Running Speed.}
We calculated the average running time of all test frames for car to measure PillarTrack's speed. PillarTrack model achieved 27 FPS on a single NVIDIA 3080 GPU, while baseline model SMAT only has 18.2 FPS. We did not introduce extra computational parameters, and we further reduced the computational overhead on the backbone while increasing the model's performance. This once again demonstrates the effectiveness of our PillarTrack. Furthermore, our PillarTrack can leverage quantization techniques\cite{lidar-ptq} to accelerate the runtime. We leave this problem for the future research.

\section{CONCLUSIONS}
In this work, we proposed PillarTrack, a pillar-based 3D single-object tracking framework that decreases computation overhead while improving performance. We propose a Pyramid-Encoding Pillar Feature Encoder (PE-PFE) and a modality-aware Transformer-based backbone to enhance the pillar representation. Extensive experiments on the KITTI and nuScenes demonstrate the superior performance of PillarTrack. Based on above designs and its efficiency, We hope our method could provide new insights for the design of 3D SOT network based on point clouds. 

\bibliographystyle{ieeetr}
\bibliography{refenrences}

\begin{thebibliography}{10}

\bibitem{fang20203d}
Z.~Fang, S.~Zhou, Y.~Cui, and S.~Scherer, ``3d-siamrpn: An end-to-end learning method for real-time 3d single object tracking using raw point cloud,'' {\em IEEE Sensors Journal}, vol.~21, no.~4, pp.~4995--5011, 2020.

\bibitem{p2b}
H.~Qi, C.~Feng, Z.~Cao, F.~Zhao, and Y.~Xiao, ``P2b: Point-to-box network for 3d object tracking in point clouds,'' in {\em Proceedings of the IEEE/CVF conference on computer vision and pattern recognition}, pp.~6329--6338, 2020.

\bibitem{ptt}
J.~Shan, S.~Zhou, Z.~Fang, and Y.~Cui, ``Ptt: Point-track-transformer module for 3d single object tracking in point clouds,'' in {\em 2021 IEEE/RSJ International Conference on Intelligent Robots and Systems (IROS)}, pp.~1310--1316, IEEE, 2021.

\bibitem{bat}
C.~Zheng, X.~Yan, J.~Gao, W.~Zhao, W.~Zhang, Z.~Li, and S.~Cui, ``Box-aware feature enhancement for single object tracking on point clouds,'' in {\em International Conference on Computer Vision}, pp.~13199--13208, 2021.

\bibitem{lttr}
Y.~Cui, Z.~Fang, J.~Shan, Z.~Gu, and S.~Zhou, ``3d object tracking with transformer,'' in {\em The British Machine Vision Conference}, 2021.

\bibitem{zhou2022pttr}
C.~Zhou, Z.~Luo, Y.~Luo, T.~Liu, L.~Pan, Z.~Cai, H.~Zhao, and S.~Lu, ``Pttr: Relational 3d point cloud object tracking with transformer,'' in {\em Computer Vision and Pattern Recognition}, pp.~8531--8540, 2022.

\bibitem{hui2022stnet}
L.~Hui, L.~Wang, L.~Tang, K.~Lan, J.~Xie, and J.~Yang, ``3d siamese transformer network for single object tracking on point clouds,'' in {\em ECCV}, 2022.

\bibitem{SiamFC}
L.~Bertinetto, J.~Valmadre, J.~F. Henriques, A.~Vedaldi, and P.~H.~S. Torr, ``Fully-convolutional siamese networks for object tracking,'' in {\em European Conference on Computer Vision (ECCV)}, pp.~850--865, 2016.

\bibitem{SiameseRPN}
B.~Li, J.~Yan, W.~Wu, Z.~Zhu, and X.~Hu, ``High performance visual tracking with siamese region proposal network,'' in {\em IEEE Conference on Computer Vision and Pattern Recognition (CVPR)}, pp.~8971--8980, 2018.

\bibitem{qi2017pointnet}
C.~R. Qi, H.~Su, K.~Mo, and L.~J. Guibas, ``Pointnet: Deep learning on point sets for 3{D} classification and segmentation,'' in {\em CVPR}, pp.~652--660, 2017.

\bibitem{qi2017pointnet++}
C.~R. Qi, L.~Yi, H.~Su, and L.~J. Guibas, ``Pointnet++: Deep hierarchical feature learning on point sets in a metric space,'' in {\em NeurIPS}, pp.~5099--5108, 2017.

\bibitem{sc3d}
S.~Giancola, J.~Zarzar, and B.~Ghanem, ``Leveraging shape completion for 3d siamese tracking,'' in {\em Proceedings of the IEEE/CVF conference on computer vision and pattern recognition}, pp.~1359--1368, 2019.

\bibitem{ptt-journal}
S.~Jiayao, S.~Zhou, Y.~Cui, and Z.~Fang, ``Real-time 3d single object tracking with transformer,'' {\em IEEE Transactions on Multimedia}, pp.~1--1, 2022.

\bibitem{shi2022pillarnet}
G.~Shi, R.~Li, and C.~Ma, ``Pillarnet: Real-time and high-performance pillar-based 3d object detection,'' in {\em ECCV}, 2022.

\bibitem{zhou2023fastpillars}
S.~Zhou, Z.~Tian, X.~Chu, X.~Zhang, B.~Zhang, X.~Lu, C.~Feng, Z.~Jie, P.~Y. Chiang, and L.~Ma, ``Fastpillars: A deployment-friendly pillar-based 3d detector,'' {\em arXiv preprint arXiv:2302.02367}, 2023.

\bibitem{smat}
Y.~Cui, J.~Shan, Z.~Gu, Z.~Li, and Z.~Fang, ``Exploiting more information in sparse point cloud for 3d single object tracking,'' {\em IEEE Robotics and Automation Letters}, vol.~7, no.~4, pp.~11926--11933, 2022.

\bibitem{pttr_journal}
Z.~Luo, C.~Zhou, L.~Pan, G.~Zhang, T.~Liu, Y.~Luo, H.~Zhao, Z.~Liu, and S.~Lu, ``Exploring point-bev fusion for 3d point cloud object tracking with transformer,'' {\em IEEE transactions on pattern analysis and machine intelligence}, 2024.

\bibitem{pointpillars}
A.~H. Lang, S.~Vora, H.~Caesar, L.~Zhou, J.~Yang, and O.~Beijbom, ``Pointpillars: Fast encoders for object detection from point clouds,'' in {\em Proceedings of the IEEE/CVF conference on computer vision and pattern recognition}, pp.~12697--12705, 2019.

\bibitem{pvt}
W.~Wang, E.~Xie, X.~Li, D.-P. Fan, K.~Song, D.~Liang, T.~Lu, P.~Luo, and L.~Shao, ``Pyramid vision transformer: A versatile backbone for dense prediction without convolutions,'' in {\em Proceedings of the IEEE/CVF international conference on computer vision}, pp.~568--578, 2021.

\bibitem{pvtV2}
W.~Wang, E.~Xie, X.~Li, D.-P. Fan, K.~Song, D.~Liang, T.~Lu, P.~Luo, and L.~Shao, ``Pvt v2: Improved baselines with pyramid vision transformer,'' {\em Computational Visual Media}, vol.~8, no.~3, pp.~415--424, 2022.

\bibitem{cui2019point}
Y.~Cui, Z.~Fang, and S.~Zhou, ``Point siamese network for person tracking using 3d point clouds,'' {\em Sensors}, vol.~20, no.~1, p.~143, 2019.

\bibitem{mmtrack}
C.~Zheng, X.~Yan, H.~Zhang, B.~Wang, S.~Cheng, S.~Cui, and Z.~Li, ``Beyond 3d siamese tracking: A motion-centric paradigm for 3d single object tracking in point clouds,'' in {\em Proceedings of the IEEE/CVF Conference on Computer Vision and Pattern Recognition}, pp.~8111--8120, 2022.

\bibitem{v2b}
L.~Hui, L.~Wang, M.~Cheng, J.~Xie, and J.~Yang, ``3d siamese voxel-to-bev tracker for sparse point clouds,'' {\em Advances in Neural Information Processing Systems}, vol.~34, pp.~28714--28727, 2021.

\bibitem{votenet}
C.~R. Qi, O.~Litany, K.~He, and L.~J. Guibas, ``Deep hough voting for 3d object detection in point clouds,'' in {\em ICCV}, pp.~9277--9286, 2019.

\bibitem{zheng2023effective}
C.~Zheng, X.~Yan, H.~Zhang, B.~Wang, S.~Cheng, S.~Cui, and Z.~Li, ``An effective motion-centric paradigm for 3d single object tracking in point clouds,'' {\em arXiv preprint arXiv:2303.12535}, 2023.

\bibitem{vaswani2017attention}
A.~Vaswani, N.~Shazeer, N.~Parmar, J.~Uszkoreit, L.~Jones, A.~N. Gomez, {\L}.~Kaiser, and I.~Polosukhin, ``Attention is all you need,'' {\em Advances in neural information processing systems}, vol.~30, 2017.

\bibitem{guo2022cmt}
Z.~Guo, Y.~Mao, W.~Zhou, M.~Wang, and H.~Li, ``Cmt: Context-matching-guided transformer for 3d tracking in point clouds,'' in {\em European Conference on Computer Vision}, pp.~95--111, Springer, 2022.

\bibitem{glt}
J.~Nie, Z.~He, Y.~Yang, M.~Gao, and J.~Zhang, ``Glt-t: Global-local transformer voting for 3d single object tracking in point clouds,'' in {\em Proceedings of the AAAI Conference on Artificial Intelligence}, vol.~37, pp.~1957--1965, 2023.

\bibitem{towards}
J.~Nie, Z.~He, X.~Lv, X.~Zhou, D.-K. Chae, and F.~Xie, ``Towards category unification of 3d single object tracking on point clouds,'' {\em arXiv preprint arXiv:2401.11204}, 2024.

\bibitem{resnet}
K.~He, X.~Zhang, S.~Ren, and J.~Sun, ``Deep residual learning for image recognition,'' in {\em Proceedings of the IEEE conference on computer vision and pattern recognition}, pp.~770--778, 2016.

\bibitem{vit}
A.~Dosovitskiy, L.~Beyer, A.~Kolesnikov, D.~Weissenborn, X.~Zhai, T.~Unterthiner, M.~Dehghani, M.~Minderer, G.~Heigold, S.~Gelly, {\em et~al.}, ``An image is worth 16x16 words: Transformers for image recognition at scale,'' {\em arXiv preprint arXiv:2010.11929}, 2020.

\bibitem{swin}
Z.~Liu, Y.~Lin, Y.~Cao, H.~Hu, Y.~Wei, Z.~Zhang, S.~Lin, and B.~Guo, ``Swin transformer: Hierarchical vision transformer using shifted windows,'' 2021.

\bibitem{zhao2021point}
H.~Zhao, L.~Jiang, J.~Jia, P.~H. Torr, and V.~Koltun, ``Point transformer,'' in {\em Proceedings of the IEEE/CVF international conference on computer vision}, pp.~16259--16268, 2021.

\bibitem{santurkar2018does}
S.~Santurkar, D.~Tsipras, A.~Ilyas, and A.~Madry, ``How does batch normalization help optimization?,'' {\em Advances in neural information processing systems}, vol.~31, 2018.

\bibitem{tian2022fully}
Z.~Tian, X.~Chu, X.~Wang, X.~Wei, and C.~Shen, ``Fully convolutional one-stage 3d object detection on lidar range images,'' {\em NeurIPS}, 2022.

\bibitem{LKUnet}
J.~Shang and S.~Zhou, ``Lk-unet: Large kernel design for 3d medical image segmentation,'' in {\em ICASSP 2024 - 2024 IEEE International Conference on Acoustics, Speech and Signal Processing (ICASSP)}, pp.~1576--1580, 2024.

\bibitem{van2008tsne}
L.~Van~der Maaten and G.~Hinton, ``Visualizing data using t-sne.,'' {\em Journal of machine learning research}, vol.~9, no.~11, 2008.

\bibitem{MLVSNet}
Z.~Wang, Q.~Xie, Y.-K. Lai, J.~Wu, K.~Long, and J.~Wang, ``Mlvsnet: Multi-level voting siamese network for 3d visual tracking,'' in {\em Proceedings of the IEEE/CVF International Conference on Computer Vision (ICCV)}, pp.~3101--3110, October 2021.

\bibitem{c2ft}
B.~Fan, K.~Wang, H.~Zhang, and J.~Tian, ``Accurate 3d single object tracker with local-to-global feature refinement,'' {\em IEEE Robotics and Automation Letters}, vol.~7, no.~4, pp.~12211--12218, 2022.

\bibitem{mlset}
Q.~Wu, C.~Sun, and J.~Wang, ``Multi-level structure-enhanced network for 3d single object tracking in sparse point clouds,'' {\em IEEE Robotics and Automation Letters}, vol.~8, no.~1, pp.~9--16, 2022.

\bibitem{kitti}
A.~Geiger, P.~Lenz, and R.~Urtasun, ``Are we ready for autonomous driving? the kitti vision benchmark suite,'' in {\em 2012 IEEE conference on computer vision and pattern recognition}, pp.~3354--3361, IEEE, 2012.

\bibitem{nuscenes}
H.~Caesar, V.~Bankiti, A.~H. Lang, S.~Vora, V.~E. Liong, Q.~Xu, A.~Krishnan, Y.~Pan, G.~Baldan, and O.~Beijbom, ``nuscenes: A multimodal dataset for autonomous driving,'' in {\em Proceedings of the IEEE/CVF conference on computer vision and pattern recognition}, pp.~11621--11631, 2020.

\bibitem{kristan2016novel}
M.~Kristan, J.~Matas, A.~Leonardis, T.~Voj{\'\i}{\v{r}}, R.~Pflugfelder, G.~Fernandez, G.~Nebehay, F.~Porikli, and L.~{\v{C}}ehovin, ``A novel performance evaluation methodology for single-target trackers,'' {\em IEEE transactions on pattern analysis and machine intelligence}, vol.~38, no.~11, pp.~2137--2155, 2016.

\bibitem{lidar-ptq}
S.~Zhou, L.~Li, X.~Zhang, B.~Zhang, S.~Bai, M.~Sun, Z.~Zhao, X.~Lu, and X.~Chu, ``Lidar-ptq: Post-training quantization for point cloud 3d object detection,'' {\em International Conference on Learning Representations (ICLR)}, 2024.

\end{thebibliography}
\end{document}